\documentclass[conference]{IEEEtran}
\IEEEoverridecommandlockouts

\usepackage[utf8]{inputenc}
\usepackage[T1]{fontenc}
\usepackage{amsmath,amssymb,amsfonts}
\usepackage{graphicx}
\usepackage{booktabs}
\usepackage{multirow}
\usepackage{array}
\usepackage{subcaption}
\usepackage{xcolor}
\usepackage{hyperref}
\usepackage{cleveref}
\usepackage{siunitx}
\usepackage{algorithm}
\usepackage{authblk}
\usepackage{algpseudocode}

\graphicspath{{figures/}}

 \newcommand{\modtp}{Fusion}

\newcommand{\cls}{\textsc{cls}}

\hypersetup{colorlinks=true, citecolor=blue!60!black,
            linkcolor=blue!60!black, urlcolor=blue!60!black}

\usepackage{amssymb}
\usepackage{fancyhdr}
\fancypagestyle{firstpage}
{
    \fancyhead[L]{\footnotesize © 2026 IEEE.  Personal use of this material is permitted.  Permission from IEEE must be obtained for all other uses, in any current or future media, including reprinting/republishing this material for advertising or promotional purposes, creating new collective works, for resale or redistribution to servers or lists, or reuse of any copyrighted component of this work in other works. This paper is accepted at the 29th Euromicro Conference Series on Digital System Design (DSD) 2026.}
    \fancyhead[R]{}
}
\begin{document}

\title{Fusion: A \textbf{\underline{F}}ramework for \textbf{\underline{U}}nified \textbf{\underline{S}}equential Token Adaptat\textbf{\underline{I}}on in Visi\textbf{\underline{O}}n Tra\textbf{\underline{N}}sformers}


\author[1]{Aravind Pradeep}
\author[2]{Samira Nazari}
\author[3,4]{Mahdi Taheri} 
\author[1]{Christian Herglotz}


\affil[1]{Brandenburg University of Technology Cottbus-Senftenberg, Germany}
\affil[2]{University of Zanjan, Iran}
\affil[3]{Humboldt University, Berlin, Germany}
\affil[4]{Tallinn University of Technology, Tallinn, Estonia}

\maketitle
\thispagestyle{firstpage}

  \begin{abstract}
  Vision Transformers achieve strong image classification accuracy but process all image regions with nearly the same computation, even when many regions are redundant or uninformative. Recent adaptive inference methods reduce this cost by selectively compressing tokens or terminating inference early, but combining these mechanisms often causes unstable intermediate representations and accuracy degradation. We introduce \modtp{}, a unified adaptive inference framework that coordinates token merging, early exiting, and token pruning through a simple staged design: tokens are merged first, confidence is evaluated next, and pruning is applied only to samples that continue inference. This ordering allows the three mechanisms to operate cooperatively rather than competitively. \modtp{} further includes lightweight routing modules that adapt compression strength to each input and support inference-time adjustment of the accuracy--latency trade-off without retraining. On ImageNet-1k with DeiT-S, \modtp{} matches or surpasses state-of-the-art adaptive ViT methods at comparable compute budgets while reducing calibration error by up to $4\times$ and inference energy by $48\%$. Experiments across ImageNet-100, CIFAR-100, and ImageNette with multiple ViT backbones demonstrate consistent transferability without dataset-specific tuning.
  \end{abstract}


\section{Introduction}
\label{sec:intro}

Vision Transformers (ViTs)~\cite{dosovitskiy2021vit,touvron2021deit}
achieve strong image classification accuracy but incur substantial computational cost due to uniform processing of all tokens across all transformer blocks. A $224\!\times\!224$ image produces 196 patch tokens that traverse all twelve layers of a DeiT-S backbone~\cite{touvron2021deit} (a 22M-parameter data-efficient Vision Transformer), regardless of whether the input image is simple or highly structured. This design leads to significant redundancy during inference.

Dynamic token processing methods address this inefficiency along three complementary axes: token pruning~\cite{rao2021dynamicvit,liang2022evit} removes uninformative tokens, token merging~\cite{bolya2023tome} aggregates redundant tokens, and early exit~\cite{yin2022avit,xu2023lgvit,teerapittayanon2016branchynet} reduces the number of transformer blocks evaluated for a given input, terminating inference once the prediction is sufficiently confident. These mechanisms are typically developed and evaluated independently, each treating the ViT backbone as a fixed computation graph.

However, combining these mechanisms does not yield additive efficiency gains. When pruning, merging, and early exit are applied simultaneously as independently trained modules, their adaptive decisions interfere through shared intermediate representations (the per-block token embeddings produced inside the transformer). On ImageNet-1k
with DeiT-S, naïve parallel composition reduces accuracy by $1.62\%,$ relative to the baseline and by $1.58\%,$ relative to a sequential composition of the same mechanisms (\Cref{tab:ablation}). The
early-exit head suffers the most: at the 8th transformer block, its standalone accuracy drops from $89.4\%$ to $47.6\%$ under parallel coupling (\Cref{tab:ablation}). While prior approaches such as ToFu~\cite{kim2024tofu} avoid this issue through joint operator design, the interaction between independently motivated mechanisms remains insufficiently understood.

This paper introduces \modtp{}, a unified framework for adaptive inference in Vision Transformers. \modtp{} coordinates token merging, early exiting, and token pruning through a staged pipeline:
merge $\rightarrow$ exit-check $\rightarrow$ prune. This simple ordering reduces cross-mechanism interference and enables stable and efficient adaptive inference. The framework includes three main components:

\begin{itemize}
    \item \textbf{Sequential adaptive inference}: merging is applied before confidence evaluation, while pruning is performed only for samples that continue inference.

    \item \textbf{Inline merge routers}: lightweight modules that dynamically predict image-specific merge ratios using the class token and token similarity supervision.

    \item \textbf{Profile-aware scaling}: an inference-time scaling strategy that supports multiple accuracy--latency trade-offs without retraining.
\end{itemize}

\modtp{} improves efficiency, calibration, and accuracy--compute trade-offs while remaining compatible with standard ViT architectures. Experiments on ImageNet-1k, ImageNet-100, CIFAR-100, and ImageNette with DeiT-S and ViT-Tiny demonstrate consistent improvements without dataset-specific tuning. The remainder of the paper is organized as follows. \Cref{sec:related} reviews related work, \Cref{sec:method} presents the proposed framework, and \Cref{sec:experiments} reports experimental results and ablation studies.

\section{Related Work}
\label{sec:related}

Adaptive inference in Vision Transformers is commonly explored through three mechanisms: token pruning, token merging, and early exit. These approaches reduce computation from different perspectives and are usually studied independently. In contrast, \modtp{} focuses on coordinating them within a unified sequential framework.

Token pruning reduces computation by removing tokens with low contribution to the final prediction. DynamicViT~\cite{rao2021dynamicvit} introduces lightweight importance predictors between transformer blocks, while EViT~\cite{liang2022evit} uses class-token attention and aggregates discarded information into a ``dustbin'' token. A-ViT~\cite{yin2022avit} formulates token reduction as adaptive halting within self-attention. Subsequent methods improve efficiency or deployment characteristics: ATS~\cite{fayyaz2022ats} replaces learned predictors with attention-based scoring, SPViT~\cite{kong2022spvit} incorporates latency-aware soft pruning, SaiT~\cite{li2022sait} adapts pruning ratios to inference budgets, and DToP~\cite{tang2023dtop} extends pruning to dense prediction tasks. Unlike these methods, \modtp{} studies pruning as part of a broader adaptive pipeline, where pruning order directly affects the stability of downstream decisions.

Token merging reduces computation by aggregating similar tokens instead of discarding them. ToMe~\cite{bolya2023tome} performs bipartite matching within each transformer block and merges highly similar token pairs using averaging, typically with fixed merge ratios. In contrast, \modtp{} predicts merge ratios dynamically for each image and transformer block using lightweight routing modules trained jointly with the backbone. Moreover, merging is applied before pruning and early exit, reducing interference between adaptive operations.

Early-exit methods reduce inference depth by terminating computation once intermediate predictions become sufficiently confident. BranchyNet~\cite{teerapittayanon2016branchynet} introduces auxiliary classifiers for early termination in convolutional networks, a strategy later adapted to Vision Transformers. LGViT~\cite{xu2023lgvit} employs SpatialPool-based exit heads, while CF-ViT~\cite{chen2023cfvit} formulates inference as a coarse-to-fine refinement process. MEViT~\cite{shen2023mevit} explores multi-exit learning for fine-grained recognition, and related NLP works such as PABEE~\cite{zhou2020pabee} and PCEE-BERT~\cite{zhang2022pceebert} investigate confidence-based and patience-based exit strategies. Unlike prior work that studies adaptive depth in isolation, \modtp{} integrates early exit into a staged token adaptation pipeline designed to preserve representation stability.

Several recent approaches combine multiple adaptive mechanisms within a single architecture. ToFu~\cite{kim2024tofu} and LTMP~\cite{bonnaerens2023ltmp} jointly design pruning and merging through unified operators, while AdaViT~\cite{meng2022adavit} learns a shared controller over layers, heads, and tokens. Slimmable Networks~\cite{yu2019slimmable} provide a related multi-profile paradigm through dynamic width scaling. Unlike these approaches, \modtp{} does not redesign adaptive mechanisms into a single coupled operator. Instead, it explicitly studies the interference arising from independently trained modules and mitigates it through sequential composition.

\section{Method}
\label{sec:method}

\begin{figure}[t]
  \centering
  \includegraphics[width=\linewidth]{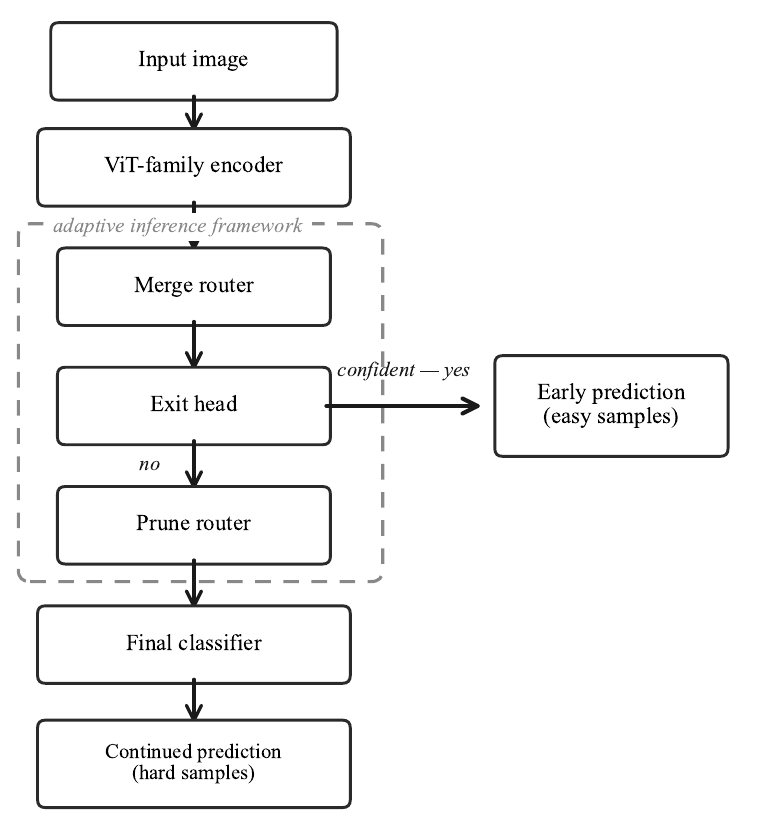}
  \caption{\modtp{} framework.}
  \label{fig:framework-flow}
\end{figure}

This section presents \modtp{}, a framework for sequential adaptive inference in Vision Transformers. \modtp{} coordinates token merging, early exit, and token pruning within a unified inference pipeline designed to minimise interference between adaptive decisions. Rather than introducing a new transformer architecture, the framework defines a lightweight routing strategy that integrates into standard ViT-family backbones. An overview is shown in Figure~\ref{fig:framework-flow}.

\subsection{Preliminaries}
\label{sec:method-prelim}

Let $f_\Theta$ denote a Vision Transformer with $L$ transformer blocks. Given an input image $\mathbf{x}\in\mathbb{R}^{3\times H\times W}$, patch embedding produces an initial token sequence
\begin{equation}
\mathbf{z}_0 \in \mathbb{R}^{(N+1)\times d},
\end{equation}
where $N$ denotes the number of patch tokens and $d$ the embedding dimension. The sequence contains one \textsc{cls} token (a learned classification vector). Each transformer block applies multi-head self-attention followed by a feed-forward network, with residual connections and layer normalisation; block $b$ computes
\begin{equation}
\mathbf{z}_b = \mathrm{Block}_b(\mathbf{z}_{b-1}),
\end{equation}
for $b\in\{1,\dots,L\}$. The final prediction is produced from the last-layer \textsc{cls} representation:

\begin{equation}
\mathbf{y} = \mathrm{head}(\mathbf{z}_L^{[\cls]}).
\end{equation}
Here $\mathrm{head}(\cdot)$ denotes the final linear classifier acting on the last-layer \textsc{cls} representation.
  \modtp{} introduces three adaptive mechanisms operating over complementary computational dimensions: token merging at
  layers $\mathcal{M}$, confidence-based early exit at layers $\mathcal{E}$, and token pruning at layers $\mathcal{P}$,
  with $\mathcal{M}, \mathcal{E}, \mathcal{P} \subseteq \{1,\dots,L\}$.

For merge layer $b\in\mathcal{M}$, a merge router predicts a merge ratio $m_b\in[0,1]$. For exit layer $b\in\mathcal{E}$, an exit head produces confidence score $c_b\in[0,1]$. For pruning layer $b\in\mathcal{P}$, a pruning router predicts token importance scores used to retain a fraction $\tau_b\in[0,1]$ of tokens.

Inference operates under profile $\theta\in\{\textsc{balanced},\textsc{power\_save}\}$, which controls the efficiency--accuracy trade-off through inference-time threshold scaling without retraining. The \textsc{balanced} setting matches the training configuration; additional profiles, such as a high-accuracy mode with weaker compression, can be obtained by adjusting the same thresholds.

\subsection{Sequential Composition of Adaptive Mechanisms}
\label{sec:method-routing}

Token merging, early exit, and token pruning reduce redundancy along distinct computational dimensions. Although complementary in principle, independently trained adaptive mechanisms interfere when executed simultaneously on shared intermediate representations.
Parallel composition substantially degrades both classification accuracy and exit reliability. The degradation is primarily caused by irreversible token removal prior to confidence estimation.

To reduce this interference, \modtp{} enforces the following execution policy:

\begin{enumerate}
    \item Token merging is applied first to compress redundant representations while preserving information through aggregation;
    \item Exit evaluation operates on merged but unpruned representations;
    \item Token pruning is deferred to later stages and applied only to samples that continue inference.
  \end{enumerate}

\begin{figure*}[!htbp]
\centering
\includegraphics[width=\linewidth]{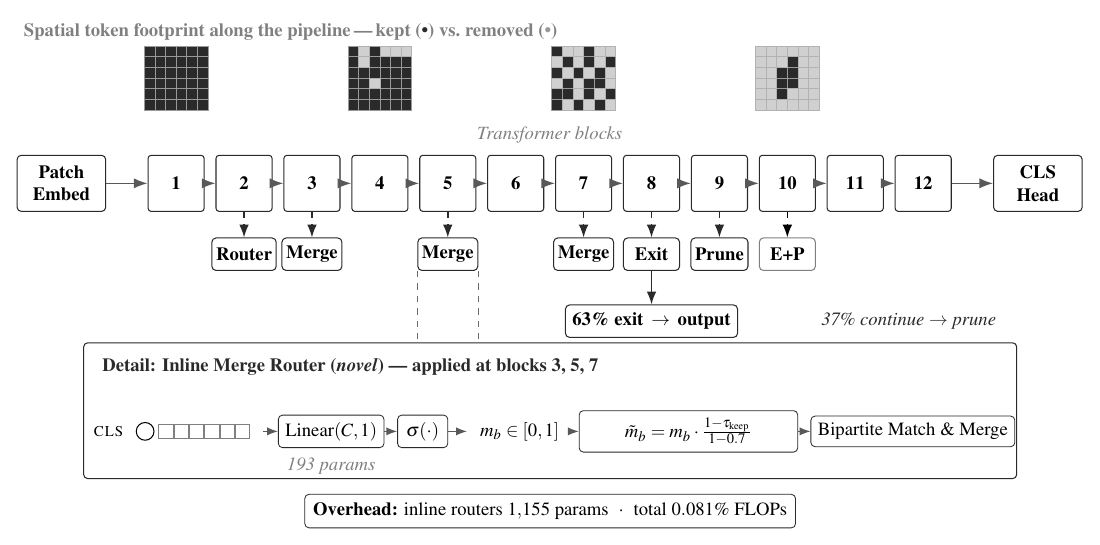}
\caption{Example DeiT-S instantiation of \modtp{}. Merge routers reduce token redundancy at intermediate layers, exit heads enable adaptive depth, and pruning routers remove low-importance tokens for samples that continue inference. The bottom panel illustrates the merge router and profile-aware scaling mechanism.}
\label{fig:architecture}
\end{figure*}

Figure~\ref{fig:architecture} illustrates a DeiT-S instantiation of the proposed sequential pipeline. The complete inference process is summarised in Algorithm~\ref{alg:inference}.

\begin{algorithm}[t]
\caption{\modtp{} inference under profile $\theta$.}
\label{alg:inference}
\begin{algorithmic}[1]
\Require Image $\mathbf{x}$, profile $\theta$
\State $\mathbf{z}_0 \gets \mathrm{PatchEmbed}(\mathbf{x})$
\For{$b=1,\dots,L$}
    \State $\mathbf{z}_b \gets \mathrm{Block}_b(\mathbf{z}_{b-1})$

    \If{$b \in \mathcal{M}$}
        \State compute $\tilde{m}_b$ using Eq.~\eqref{eq:profile-scaling}
        \State $\mathbf{z}_b \gets \mathrm{Merge}(\mathbf{z}_b,\tilde{m}_b)$
    \EndIf

    \If{$b \in \mathcal{E}$}
        \State compute $c_b$ using Eq.~\eqref{eq:exit-rule}
        \If{$c_b \geq \delta_b(\theta)$}
            \State \Return $\mathrm{head}_b(\mathbf{z}_b)$
        \EndIf
    \EndIf

    \If{$b \in \mathcal{P}$}
        \State $\mathbf{z}_b \gets \mathrm{Prune}(\mathbf{z}_b,\tau_b)$
    \EndIf
\EndFor

\State \Return $\mathrm{head}(\mathbf{z}_L^{[\cls]})$
\end{algorithmic}
\end{algorithm}

\subsection{Adaptive Token Merging}
\label{sec:method-merge}

Token merging reduces computation by aggregating similar token representations. Unlike fixed-rate merging approaches~\cite{bolya2023tome}, \modtp{} predicts merge intensity dynamically for each input and transformer layer.

For merge layer $b\in\mathcal{M}$, the router is a single linear projection $g_b:\mathbb{R}^d\rightarrow\mathbb{R}$ with weight
vector $\mathbf{w}_b\in\mathbb{R}^d$ and bias $\beta_b\in\mathbb{R}$. It receives the \textsc{cls} representation and predicts a
per-block \emph{merge ratio} $m_b\in[0,1]$, the fraction of token pairs to merge at layer $b$:
\begin{equation}
  m_b = \sigma\!\left(\mathbf{w}_b^\top\,\mathbf{z}_b^{[\cls]} + \beta_b\right),
\end{equation}
where $\sigma(\cdot)$ denotes the sigmoid function. Each router adds only $d+1$ parameters per merge layer (1{,}155 total for DeiT-S across three merge layers).
To support multiple operating profiles using a single checkpoint, merge ratios are rescaled during inference according to the target computational budget:

\begin{equation}
\tilde{m}_b
=
m_b
\cdot
\frac{1-\tau_{\mathrm{tgt}}(\theta)}
{1-\tau_{\mathrm{tgt}}(\theta_{\mathrm{ref}})},
\label{eq:profile-scaling}
\end{equation}
where $\tau_{\mathrm{tgt}}(\theta)$ denotes the target keep ratio associated with profile $\theta$, and $\theta_{\mathrm{ref}}$ denotes the reference training profile.

Given $\tilde{m}_b$ and the current token count $N_b$ at layer $b$ (which may already be reduced by prior merging), bipartite token matching from
ToMe~\cite{bolya2023tome} is applied: tokens are split into two halves and each token in one half is paired with its most cosine-similar counterpart in
the other, yielding $N_b/2$ candidate pairs. The merge ratio $\tilde{m}_b\in[0,1]$ then selects how many to merge --- the top $\lfloor \tilde{m}_b N_b/2
\rfloor$ pairs are averaged into single tokens, reducing the layer's token count by that amount. This step corresponds to Algorithm~\ref{alg:inference},
lines 4--7.
\subsection{Confidence-Based Early Exit}
\label{sec:method-exit}

Early exit reduces effective inference depth by terminating computation once intermediate representations become sufficiently confident.

For exit layer $b\in\mathcal{E}$, an auxiliary classifier produces

\begin{equation}
\hat{p}_b
=
\mathrm{softmax}
\big(
\mathrm{head}_b(\mathbf{z}_b)
\big).
\end{equation}
 Here $\mathrm{head}_b(\cdot)$ is the auxiliary classifier at exit layer $b$.
Inference terminates at block $b$ if

\begin{equation}
c_b
=
\max_k \hat{p}_b^{(k)}
\geq
\delta_b(\theta),
\label{eq:exit-rule}
\end{equation}

 where $\delta_b(\theta)$ is a profile-dependent confidence threshold selected on a held-out validation set to meet the target FLOPs budget for each
  profile, without retraining. Exit evaluation precedes pruning so that confidence estimation operates on unreduced token representations.
This step corresponds to
Algorithm~\ref{alg:inference}, lines 8--13.

\subsection{Progressive Token Pruning}
\label{sec:method-prune}

Samples that do not exit early continue to late-stage token pruning. At deeper transformer layers, token importance becomes increasingly concentrated, enabling aggressive reduction with limited impact on prediction quality.

For pruning layer $b\in\mathcal{P}$, a pruning router predicts token importance scores

\begin{equation}
s_i = h_b(\mathbf{z}_{b,i}),
\end{equation}

  where $h_b$ is a lightweight scoring head and $\mathbf{z}_{b,i}$ denotes the representation of token $i$ at layer $b$. The top $\lfloor \tau_b N_b
  \rfloor$ tokens are retained, while the remaining tokens are discarded. The keep ratio $\tau_b\in[0,1]$ is a per-layer hyperparameter, selected on a
  held-out validation set jointly with the exit thresholds $\delta_b$ to meet the target FLOPs budget for each profile (Algorithm~\ref{alg:inference}, lines 14--16).





\subsection{Training Objective}
\label{sec:method-training}

All adaptive components are trained jointly using a teacher-distilled multi-objective loss~\cite{hinton2015distilling}. Let $\mathbf{y}^*$ denote the
ground-truth label, $\mathbf{p}$ the model's output class distribution, and $\mathbf{p}_T$ the temperature-scaled teacher distribution. The overall
objective combines a cross-entropy term $\mathcal{L}_{\mathrm{CE}}$ and a Kullback--Leibler divergence term $\mathcal{L}_{\mathrm{KL}}$:

\begin{align}
\mathcal{L}
=
&
\mathcal{L}_{\mathrm{CE}}(\mathbf{y},\mathbf{y}^*)
+
\alpha T^2
\mathcal{L}_{\mathrm{KL}}(\mathbf{p},\mathbf{p}_T)
\nonumber\\
&
+
\lambda_{\mathrm{exit}}
\sum_{b\in\mathcal{E}}
\Big[
\mathcal{L}_{\mathrm{CE}}(\mathbf{y}_b,\mathbf{y}^*)
+
T^2
\mathcal{L}_{\mathrm{KL}}(\mathbf{p}_b,\mathbf{p}_T)
\Big]
\nonumber\\
&
+
\lambda_{\mathrm{aux}}
\mathcal{L}_{\mathrm{router}}
+
\lambda_{\mathrm{div}}
\mathcal{L}_{\mathrm{div}}
+
\lambda_{\mathrm{budget}}
\mathcal{L}_{\mathrm{budget}}.
\label{eq:loss}
\end{align}

Here $\alpha$ is the distillation weight, $T$ the
temperature, $\mathbf{p}_b$ and $\mathbf{y}_b$
the exit-head prediction and target at layer $b$,
and $\lambda_\mathrm{exit}$,
$\lambda_\mathrm{aux}$, $\lambda_\mathrm{div}$,
 $\lambda_\mathrm{budget}$ are loss coefficients; a full parameter summary appears in \Cref{tab:params}.
$\mathcal{L}_{\mathrm{router}}$ is the MSE between the predicted merge ratio $m_b$ and per-block token-similarity statistics; $\mathcal{L}_{\mathrm{div}}$ is a routing-diversity penalty preventing $m_b$ from collapsing to a constant; and $\mathcal{L}_{\mathrm{budget}}$ is the squared deviation of mean $m_b$ from the target keep ratio.

Training proceeds in three stages: router warm-up, exit-head pretraining, and joint fine-tuning. This staged optimisation stabilises convergence and mitigates premature routing collapse.

\subsection{Sequential Inference Procedure}
\label{sec:method-inference}
During inference, computation dynamically adapts to input difficulty. \modtp{} places two exit heads (at blocks 8 and 10 in our DeiT-S configuration); easy samples terminate at the earlier head after token merging, whereas difficult samples continue through later transformer stages with progressive pruning applied.

Algorithm~\ref{alg:inference} summarises the complete inference process. The three conditional blocks correspond to the merge, exit, and prune operators defined in \Cref{sec:method-merge,sec:method-exit,sec:method-prune},
  respectively.

\section{Experimental Results}
\label{sec:experiments}

Experiments evaluate four questions: (i) whether \modtp{} improves the accuracy--efficiency trade-off relative to existing adaptive ViTs, (ii) whether the framework transfers across datasets and backbones, (iii) which design choices are responsible for the observed gains, and (iv) how the framework behaves beyond FLOPs and accuracy, including energy, calibration, and latency. \Cref{sec:exp-imagenet1k,sec:exp-fair} address~(i);
\Cref{sec:exp-cross,sec:exp-multi-backbone} address~(ii); \Cref{sec:exp-ablation} addresses~(iii); and \Cref{sec:exp-router,sec:exp-energy,sec:exp-cost}
  address~(iv).

\subsection{Experimental Setup}
\label{sec:exp-setup}

\paragraph{Datasets}
All experiments are single-label image-classification benchmarks: $1000$ classes on ImageNet-1k~\cite{deng2009imagenet}, $100$ classes on ImageNet-100,
$100$ classes on CIFAR-100, and $10$ classes on ImageNette. All inputs are resized to $224\times224$ and normalised using the default ViT preprocessing
pipeline.

  \begin{table}[t]
  \centering
  \caption{Parameters of \modtp{} on DeiT-S.}
  \label{tab:params}
  \footnotesize
  \setlength{\tabcolsep}{4pt}
  \begin{tabular}{lll}
  \toprule
  Parameter & Determination & Value (DeiT-S, balanced) \\
  \midrule
  Backbone weights & Trained  & $\sim$22M \\
  Inline merge routers $g_b$ & Trained  & $1{,}155$ ($3{\times}(d{+}1)$) \\
  Exit heads & Trained  & $\sim$2.1M total \\
  Pruning router $h_b$ & Trained & lightweight head \\
  \midrule
  Layer sets $\mathcal{M},\mathcal{E},\mathcal{P}$ & Heuristic  & $\{3,5,7\}, \{8,10\}, \{9,10\}$ \\
  Reference profile $\theta_{\mathrm{ref}}$ & Fixed & \textsc{balanced} \\
  Exit thresholds $\delta_b(\theta)$ & Validation-tuned & $(0.8, 0.95)$ \\
  Keep ratios $\tau_b(\theta)$ & Validation-tuned & $\tau_{\mathrm{tgt}}=0.7$ \\
  \midrule
  KD $\alpha, T$ & Fixed & $0.5, 4.0$ \\
  $\lambda_{\mathrm{exit}}, \lambda_{\mathrm{aux}}, \lambda_{\mathrm{div}}, \lambda_{\mathrm{budget}}$ & Fixed & $0.5, 0.1, 0.01, 0.1$ \\
  \bottomrule
  \end{tabular}
  \end{table}

\paragraph{Backbones}
Results are reported for DeiT-S~\cite{touvron2021deit} and ViT-Tiny~\cite{dosovitskiy2021vit}, initialised from ImageNet-1k pretrained checkpoints provided by the \texttt{timm} library~\cite{rw2019timm} (PyTorch Image Models). The merge layer set $\mathcal{M}$, exit layer set $\mathcal{E}$, and pruning layer set $\mathcal{P}$ are specified in \Cref{tab:params}.

\paragraph{Baselines}
Comparisons include three categories of adaptive ViT methods: token-pruning baselines DynamicViT~\cite{rao2021dynamicvit} and EViT~\cite{liang2022evit}; the adaptive-halting baseline A-ViT~\cite{yin2022avit}; and the token-merging baseline ToMe~\cite{bolya2023tome}. Published ImageNet-1k numbers are reported where available, together with same-session comparisons under identical hardware and preprocessing.

  \paragraph{Metrics}
We report Top-1 accuracy on each test set; FLOPs reduction as $1-\mathrm{FLOPs}/\mathrm{FLOPs}_{\mathrm{baseline}}$, computed analytically (static per-input); throughput and latency averaged over $100$ timed runs after $10$ warm-up iterations on RTX~3050 (dynamic, depends on per-image early-exit decisions); energy per inference integrated from GPU power via \texttt{nvmlDeviceGetPowerUsage} at $100$\,Hz; and Expected Calibration Error $\mathrm{ECE} = \sum_b (n_b/N)\,|\mathrm{acc}(b) - \mathrm{conf}(b)|$ over $15$ confidence bins, where $n_b$ is the count in bin $b$.

\paragraph{Profiles}
Unless stated otherwise, \modtp{} refers to the \textsc{balanced} profile, which matches the training configuration. The \textsc{power\_save} profile uses
tighter exit thresholds $\delta_b(\theta)$ and a lower target keep ratio $\tau_{\mathrm{tgt}}(\theta)$ via the profile-scaling rule (Eq.~\eqref{eq:profile-scaling}) at inference time, without retraining; both profiles were chosen on a held-out validation set to span the practical
accuracy--FLOPs operating range.

\subsection{Main Results on ImageNet-1k}
\label{sec:exp-imagenet1k}
\Cref{tab:imagenet1k-main} compares \modtp{} against representative adaptive ViT baselines on ImageNet-1k with DeiT-S. The \textsc{balanced} profile preserves baseline accuracy within $0.09\,$pp while reducing FLOPs by $31.8\%$. At comparable or larger FLOPs reductions it exceeds DynamicViT by $+0.46\,$pp, EViT by $+0.26\,$pp, ToMe ($r{=}13$) by $+0.36\,$pp, and A-ViT by $+1.16\,$pp. The \textsc{power\_save} profile reaches $40.1\%$ FLOPs
reduction with a $0.70\,$pp accuracy drop, matching DynamicViT's accuracy while saving $3.3\,$pp more FLOPs.

\begin{table}[t]
\centering
\caption{ImageNet-1k results with DeiT-S. Both \modtp{} profiles are obtained from a single trained checkpoint.}
\label{tab:imagenet1k-main}
\footnotesize
\setlength{\tabcolsep}{4pt}
\begin{tabular}{lcc}
\toprule
Method & Acc@1 & FLOPs $\downarrow$ \\
\midrule
DeiT-S & 79.85 & 0.0\% \\
\midrule
DynamicViT & 79.30 & 36.8\% \\
EViT & 79.50 & 35.0\% \\
A-ViT & 78.60 & 26.0\% \\
ToMe ($r{=}13$) & 79.40 & 35.0\% \\
\midrule
\modtp{} \textsc{(balanced)} & \textbf{79.76} & 31.8\% \\
\modtp{} \textsc{(power\_save)} & 79.15 & \textbf{40.1\%} \\
\bottomrule
\end{tabular}
\end{table}

\paragraph{Single-checkpoint operating range.}
Unlike prior methods that require a separate trained checkpoint for each compute--accuracy target, \modtp{} covers a broad range of operating points from one trained checkpoint via inference-time threshold scaling alone. Each \emph{profile} (e.g., \textsc{balanced}, \textsc{power\_save}) selects a specific operating point on the accuracy--FLOPs curve by adjusting the merge/exit/prune thresholds at inference time. \Cref{tab:iso-flops-pareto} shows that this single checkpoint reaches the operating points of EViT and DynamicViT while maintaining competitive or higher accuracy.

\begin{table}[t]
\centering
\caption{Apples-to-apples comparison with published baselines (DynamicViT, EViT, ToMe), all trained under the same $30$-epoch budget.}
\label{tab:iso-flops-pareto}
\footnotesize
\setlength{\tabcolsep}{4pt}
\begin{tabular}{llcc}
\toprule
Method & Operating point & Acc@1 & FLOPs $\downarrow$ \\
\midrule
DeiT-S & --- & 79.85 & 0.0\% \\
\midrule
\modtp{} & balanced & \textbf{79.62} & 31.9\% \\
\modtp{} & iso-EViT & 79.49 & 34.1\% \\
\modtp{} & iso-DynamicViT & \textbf{79.49} & 36.9\% \\
\modtp{} & power\_save & 79.25 & 39.7\% \\
\midrule
EViT & reference & 79.50 & 33.0\% \\
ToMe ($r{=}13$) & reference & 79.40 & 35.0\% \\
DynamicViT & reference & 79.30 & 36.8\% \\
\bottomrule
\end{tabular}
\end{table}

\begin{figure}[t]
\centering
\includegraphics[width=\linewidth]{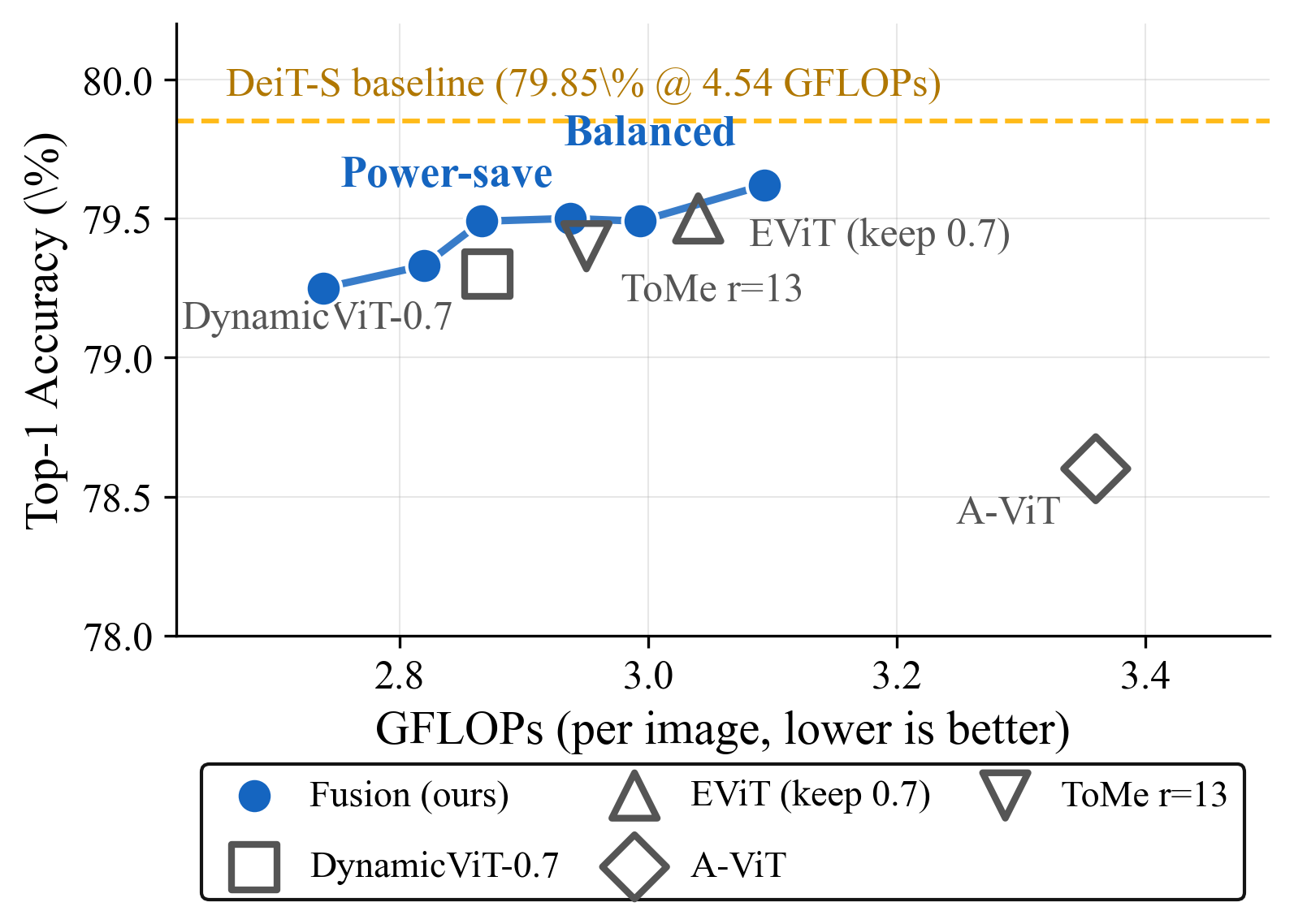}
\caption{Accuracy versus GFLOPs on ImageNet-1k with DeiT-S. \modtp{} forms the upper-left Pareto frontier across multiple operating points from a single checkpoint.}
\label{fig:pareto}
\end{figure}

The same checkpoint spans the $31$--$40\%$ FLOPs regime with only $0.37\,$pp accuracy variation, demonstrating stable multi-profile behaviour without retraining. \Cref{fig:pareto} visualises this Pareto frontier against the published baselines: \modtp{} sits on the upper-left, achieving higher accuracy at every matched FLOPs budget.

\subsection{Same-Session Comparison}
\label{sec:exp-fair}

To eliminate differences in hardware setup and preprocessing pipelines across published baselines, DynamicViT, EViT, ToMe, and \modtp{} are evaluated under identical conditions. Results are shown in \Cref{tab:fair-comparison}.

\begin{table}[t]
\centering
\caption{Same-session comparison on ImageNet-1k under identical hardware and preprocessing.}
\label{tab:fair-comparison}
\small
\begin{tabular}{lccc}
\toprule
Method & Acc@1 & FLOPs $\downarrow$ & Energy (mJ/img) \\
\midrule
DeiT-S & 79.85 & 0.0\% & 313.6 \\
DynamicViT & 79.23 & 36.8\% & 245.4 \\
EViT & 79.04 & 36.8\% & 246.1 \\
ToMe & 79.04 & 36.8\% & 243.1 \\
\midrule
\shortstack[l]{\modtp{} \\ \textsc{(balanced)}} & \textbf{79.59} & 32.1\% & 255.8 \\
\shortstack[l]{\modtp{} \\ \textsc{(power\_save)}} & 79.16 & 39.8\% & \textbf{163.4} \\
\bottomrule
\end{tabular}
\end{table}

The \textsc{balanced} profile achieves the highest accuracy among all adaptive baselines. The \textsc{power\_save} profile reduces inference energy by $47.9\%$, substantially exceeding the savings achieved by single-axis adaptive methods.

\subsection{Cross-Dataset Generalisation}
\label{sec:exp-cross}

\Cref{tab:cross-dataset} evaluates transfer across datasets and backbones using a shared training recipe. \modtp{} matches or exceeds baseline accuracy on all smaller datasets while maintaining substantial FLOPs reduction.

\begin{table}[t]
\centering
\caption{Cross-dataset evaluation. The training pipeline is unchanged across datasets.}
\label{tab:cross-dataset}
\small
\begin{tabular}{lcccc}
\toprule
Dataset & Backbone & Baseline & \modtp{} & FLOPs $\downarrow$ \\
\midrule
ImageNette & ViT-Tiny & 95.24 & \textbf{96.61} & 26.6\% \\
CIFAR-100 & ViT-Tiny & 85.08 & \textbf{86.50} & 37.0\% \\
ImageNet-100 & ViT-Tiny & 88.60 & \textbf{88.62} & 33.2\% \\
ImageNet-1k & DeiT-S & 79.85 & 79.76 & 31.8\% \\
\bottomrule
\end{tabular}
\end{table}

The gains on smaller datasets suggest that the auxiliary supervision introduced by multi-exit training and distillation acts as an implicit regulariser.

\subsection{Multi-Backbone Evaluation}
\label{sec:exp-multi-backbone}

The framework transfers across different transformer sizes (from $5$M-parameter ViT-Tiny to $86$M-parameter DeiT-B) without architectural modification.

\paragraph{ViT-Tiny.}
\Cref{tab:vit-tiny} reports ImageNet-1k results on ViT-Tiny. The balanced profile reaches $71.51\%$ at $31.3\%$ FLOPs reduction --- a $3.95\,$pp drop from the $75.46\%$ unmodified-backbone baseline. This drop is larger than the DeiT-S one, but \modtp{} still exceeds DynamicViT-Tiny\footnote{Our in-house training of DynamicViT on the ViT-Tiny backbone, since the published DynamicViT uses DeiT-S; this ensures a same-backbone comparison.} ($71.15\%$ at $\approx 37\%$ FLOPs reduction) by $+0.36\,$pp at lower FLOPs reduction, while using fewer joint fine-tuning epochs. The bottom row of \Cref{tab:vit-tiny}  reports a parallel-composition variant (pruning at $\{3,6,9\}$ before exit at $\{8,10\}$) as an interference reference: accuracy collapses to $51.44\%$
($-20.07\,$pp vs.\ the sequential balanced profile), confirming on this smaller backbone the same interference effect measured on DeiT-S in \Cref{tab:ablation}.
  \begin{table}[t]
  \centering
  \caption{ImageNet-1k with ViT-Tiny backbone.}
  \label{tab:vit-tiny}
  \footnotesize
  \setlength{\tabcolsep}{4pt}
  \begin{tabular}{lcc}
  \toprule
  Method & Acc@1 & FLOPs $\downarrow$ \\
  \midrule
  ViT-Tiny & 75.46 & 0.0\% \\
 DynamicViT-Tiny & 71.15 & $\approx 37\%$ \\
  \midrule
  \modtp{} \textsc{(balanced)} & \textbf{71.51} & 31.3\% \\
  \modtp{} \textsc{(power\_save)} & 70.27 & \textbf{41.2\%} \\
  \midrule
  \modtp{} parallel (interference ref.) & 51.44 & 46.9\% \\
  \bottomrule
  \end{tabular}
  \end{table}
  
\paragraph{DeiT-B.} Repeating the same three-phase training schedule (see \Cref{sec:method-training}) on the larger $86$M-parameter DeiT-B backbone, \modtp{} reaches $81.28\%$ at $32.2\%$ FLOPs reduction --- a $0.70\,$pp drop from the $81.98\%$ unmodified-backbone baseline. The \textsc{power\_save} profile reaches $40.8\%$ FLOPs reduction at a $2.08\,$pp accuracy drop. This trained variant outperforms a training-free baseline on the same backbone: zero-shot token
merging (ToMe with $r{=}8$) yields only $81.17\%$ at $37.7\%$ FLOPs reduction --- less accuracy preserved at more FLOPs reduction.

\subsection{Ablation Study}
\label{sec:exp-ablation}

\Cref{tab:ablation} isolates the contribution of each design component.

\begin{table}[t]
\centering
\caption{Ablation study on ImageNet-1k with DeiT-S.}
\label{tab:ablation}
\small
\begin{tabular}{lccc}
\toprule
Configuration & Acc@1 & Exit-8 & FLOPs $\downarrow$ \\
\midrule
DeiT-S & 79.85 & --- & 0.0\% \\
\midrule
Pruning only & 79.83 & --- & 7.0\% \\
Merging only & 79.72 & --- & 10.0\% \\
Early exit only & 79.48 & 89.4 & 21.0\% \\
\midrule
Parallel prune + exit & 78.23 & 47.6 & 41.8\% \\
Sequential exit $\rightarrow$ prune & 79.81 & 89.4 & 17.9\% \\
\midrule
 Full sequential (cascade merge) & 79.73 & 89.4 & 25.5\% \\
\midrule
\quad + inline merge routers    & 79.20 & 88.7 & 32.5\% \\
\quad + profile-aware scaling   & 79.13 & 88.5 &
  \textbf{39.8\%} \\

\bottomrule
\end{tabular}
\end{table}
\Cref{tab:ablation} is grouped into three blocks. \textbf{Single mechanisms} (rows 2--4) bound the per-axis cost: pruning, merging, and early exit each contribute small accuracy drops ($\leq 0.37\,$pp) for $7$--$21\%$ FLOPs savings. \textbf{Composition} (rows 5--6) reveals the interference effect:
applying pruning and early exit in \emph{parallel} drops accuracy by $1.62\,$pp and collapses the block-$8$ exit head from $89.4\%$ to $47.6\%$ standalone
accuracy, whereas the \emph{sequential} ordering (exit-check before prune) recovers both --- a $1.58\,$pp interference penalty avoided. \textbf{Full
pipeline} (rows 7--9) layers cascade merging at $\{3,5,7\}$, learnable inline merge routers, and profile-aware scaling of the merge ratios $\tilde m_b$
and exit thresholds $\delta_b$ (Eq.~\eqref{eq:profile-scaling}) on top of the sequential composition; each addition trades $\sim$0.1--$0.5\,$pp accuracy
  for $\sim$7\,pp more FLOPs reduction, reaching $39.8\%$ FLOPs at only $0.07\,$pp accuracy cost on top of the full sequential baseline.
\subsection{Routing Behaviour}
\label{sec:exp-router}

\begin{figure}[t]
\centering
\includegraphics[width=\linewidth]{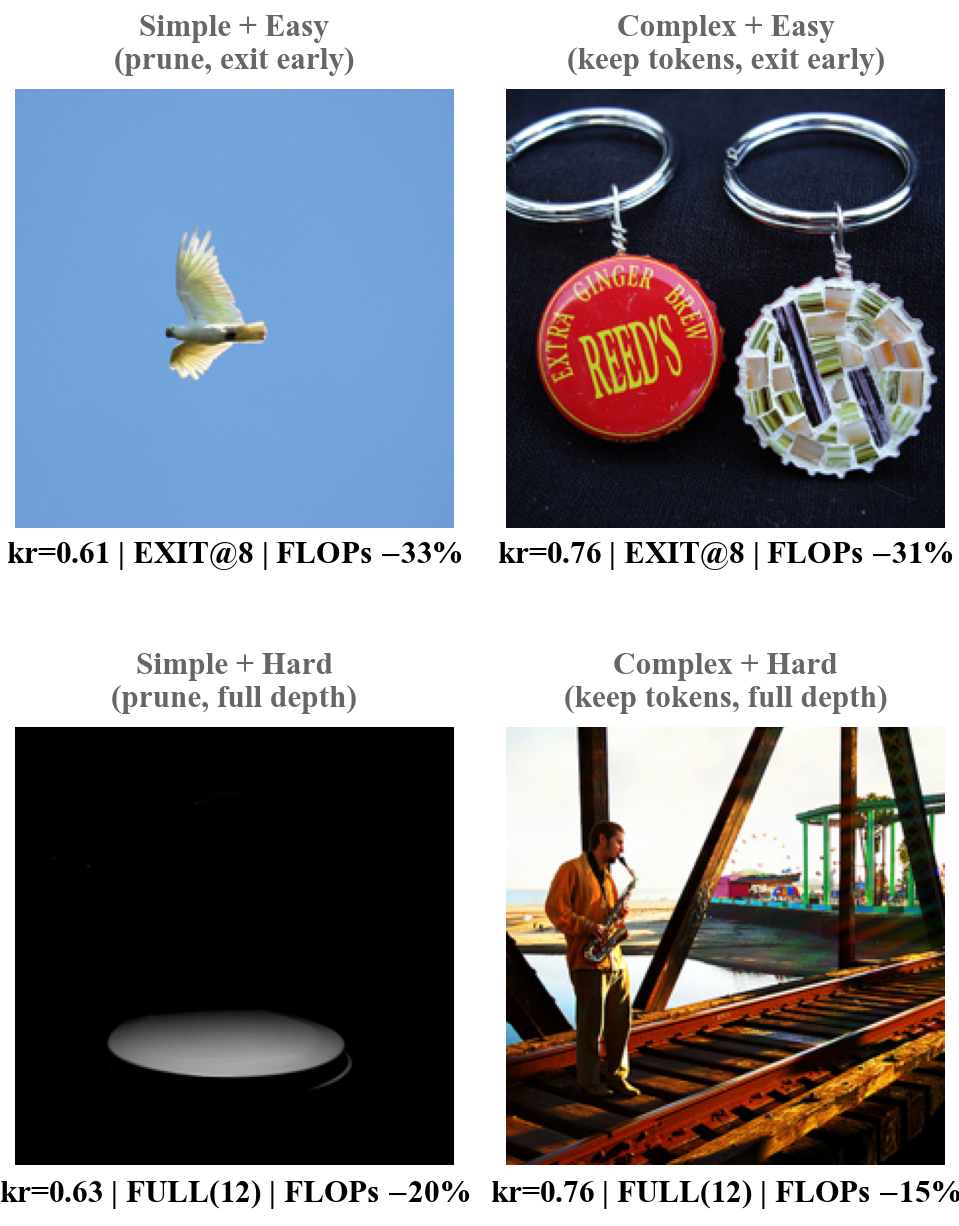}
\caption{Per-image routing behaviour on ImageNet-1k validation samples.}
\label{fig:routing}
\end{figure}

\Cref{fig:routing} visualises routing behaviour on ImageNet-1k samples. Merge ratios increase for visually redundant backgrounds and decrease for structurally complex scenes. Exit confidence is similarly correlated with semantic ambiguity.

The router output has standard deviation $0.015$ around the profile target $\tau_{\text{tgt}}=0.70$. A non-zero spread is the necessary signature of
input-conditional routing: a router collapsed to a fixed schedule would yield $\text{std}=0$ by construction. The per-image merge-ratio variation visible in \Cref{fig:routing} reflects this signal.

\subsection{Energy, Calibration, and Latency}
\label{sec:exp-energy}

\begin{figure}[t]
  \centering
  \includegraphics[width=\linewidth]{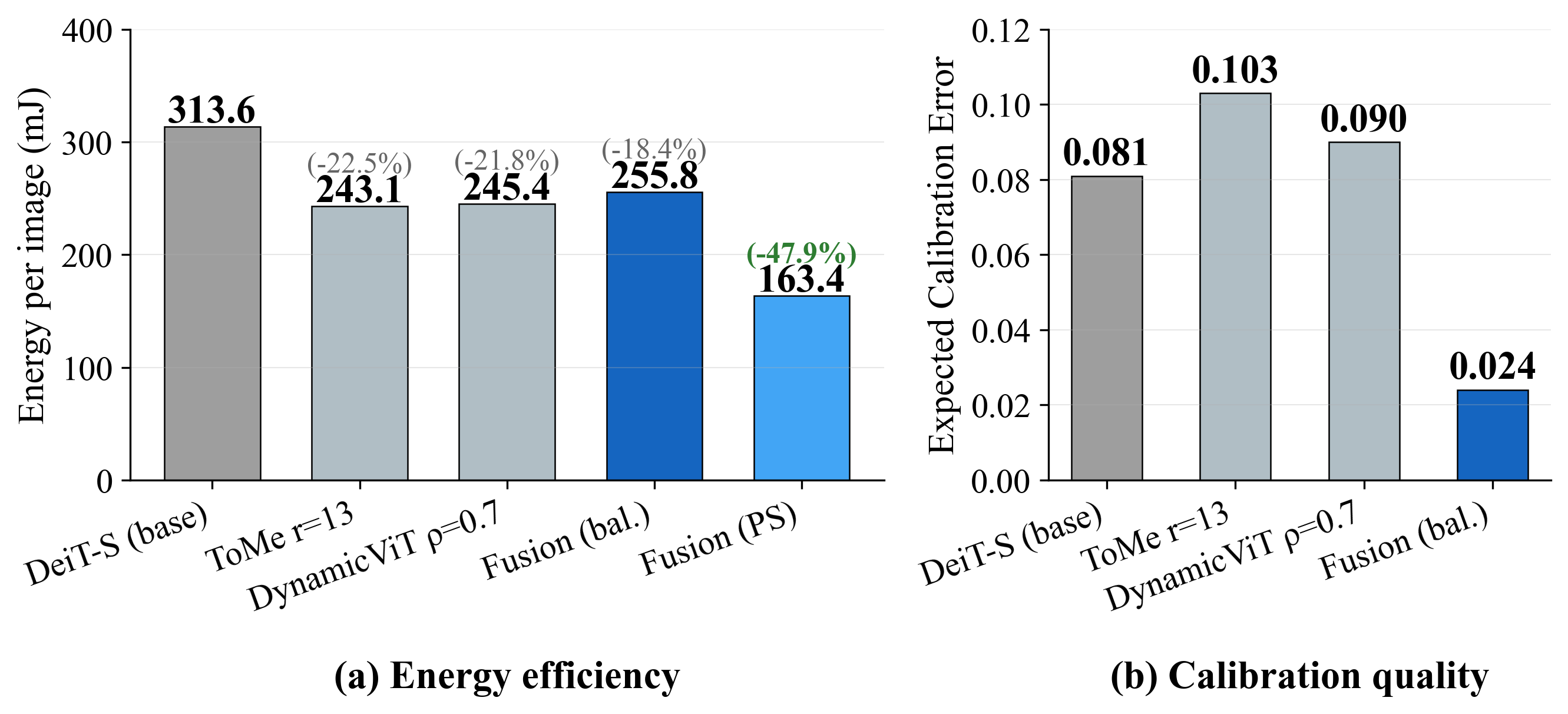}
  \caption{Energy consumption (a) and Expected Calibration Error (b) on ImageNet-1k --- lower is better on both axes.}
  \label{fig:energy-ece}
  \end{figure}

\Cref{fig:energy-ece}(a) reports energy consumption per inference (lower is better). The \textsc{power\_save} profile reduces inference energy by $47.9\%$ relative to the DeiT-S baseline --- more than $2\times$ the savings of any single-axis adaptive
baseline (DynamicViT $21.8\%$, EViT $21.5\%$, ToMe $22.5\%$). This exceeds the FLOPs reduction because samples that exit early bypass entire transformer blocks rather than just processing fewer tokens, avoiding per-block fixed costs (kernel launches, attention setup, memory traffic) that the analytical FLOPs count does not attribute to inference.

\modtp{} also improves calibration substantially. \Cref{fig:energy-ece}(b) reports Expected Calibration Error (lower is better):
\modtp{} reaches $0.024$, $3.4\times$ better than the DeiT-S baseline ($0.081$) and roughly $4\times$ better than DynamicViT
($0.090$) and ToMe ($0.103$). This improvement is consistent with the auxiliary supervision provided by the early-exit heads during joint fine-tuning: each exit head matches the teacher logits at its own depth, encouraging the backbone to produce
well-calibrated predictions at multiple intermediate layers, not only at the final classifier.

\begin{figure*}
\centering
\includegraphics[width=\linewidth]{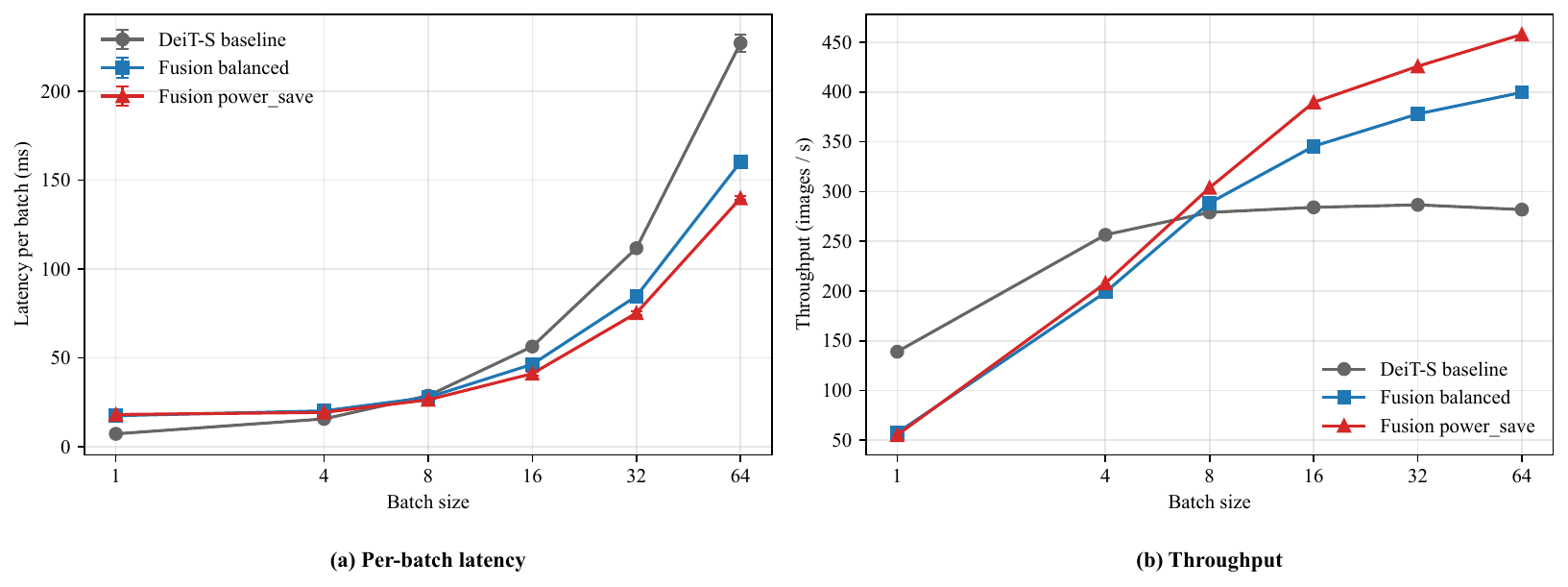}
\caption{Latency and throughput across batch sizes.}
\label{fig:latency-sweep}
\end{figure*}

Latency and throughput results are shown in \Cref{fig:latency-sweep}. The adaptive overhead dominates at batch size $1$, but throughput improves substantially once routing costs are amortised. The \textsc{power\_save} profile reaches $1.62\times$ the baseline throughput at batch size $64$.

\subsection{Parameter and Memory Overhead}
\label{sec:exp-cost}

The framework introduces minimal overhead relative to the backbone. Inline merge routers add only
$1{,}155$ parameters on DeiT-S, while the two exit heads contribute approximately $2.1$M additional
parameters (a SpatialPool head at block~8 and a CLS-only head at block~10, each a 2-layer MLP). Activation memory remains effectively unchanged relative to the baseline backbone.

\section{Conclusion}
\label{sec:conclusion}

This paper introduced \modtp{}, a unified framework for sequential multi-mechanism adaptation in Vision Transformers. Rather than treating token merging, early exit, and token pruning as independent operations, \modtp{} organises them into a sequential inference pipeline that reduces cross-mechanism interference while preserving representation quality. The framework combines lightweight merge routing, confidence-based adaptive depth, and profile-aware inference scaling within a backbone-agnostic formulation compatible with standard ViT architectures. A single trained checkpoint matches or exceeds the operating points of DynamicViT, EViT, and ToMe on ImageNet-1k while supporting multiple efficiency--accuracy trade-offs without retraining. More broadly, the results suggest that adaptive mechanisms in Vision Transformers are fundamentally interaction-dependent: once ordered correctly, multiple adaptive axes become complementary rather than destructive.
\section*{Acknowledgements}
\small
This work was supported in part by the Estonian Research Council grant PUT PRG1467 "CRASHLESS“, EU Grant Project 101160182 “TAICHIP“, by the Deutsche Forschungsgemeinschaft (DFG, German Research Foundation) – Project-ID "458578717", and by the Federal Ministry of Research, Technology and Space of Germany (BMFTR) for supporting Edge-Cloud AI for DIstributed Sensing and COmputing (AI-DISCO) project (Project-ID "16ME1127")
\bibliographystyle{IEEEtran}
\bibliography{bib_dsd}

\end{document}